\PassOptionsToPackage{disabled}{doclicense}
\documentclass[sigconf]{acmart}

\makeatletter
\def\@ACM@copyright@check@cc{}
\makeatother
\AtBeginDocument{%
  }

\usepackage[ruled,vlined,linesnumbered]{algorithm2e}

\copyrightyear{2026}
\acmYear{2026}
\setcopyright{cc}
\setcctype{by-nc-nd}
\acmConference[WSDM Companion '26]{The Nineteenth ACM International Conference on Web Search and Data Mining}{February 22--26, 2026}{Boise, ID, USA}
\acmBooktitle{The Nineteenth ACM International Conference on Web Search and Data Mining (WSDM Companion '26), February 22--26, 2026, Boise, ID, USA}
\acmPrice{}
\acmDOI{10.1145/3779211.3793166}
\acmISBN{979-8-4007-2358-2/2026/02}

\begin{document}

\title{SAGE: Scalable Automatic Gating Ensemble for Confident Negative Harvesting in Fraud Detection}

\author{Sudheer Tubati}
\email{sudheet@amazon.com}
\orcid{}
\affiliation{%
  \institution{Amazon Music}
  \city{Seattle}
  %\state{Washington}
  \country{USA}
}

\author{Amit Goyal}
\email{goyalam@amazon.com}
\orcid{}
\affiliation{%
  \institution{Amazon Music}
  \city{San Francisco}
  \country{USA}
}

%\author{Anonymous Authors}
%% \renewcommand{\shortauthors}{Trovato et al.}

\begin{abstract}
Music streaming fraud, where bad actors artificially inflate stream counts to manipulate chart rankings and royalty payments, poses a significant threat to streaming services and legitimate content creators. Traditional fraud detection approaches struggle with a critical challenge: many legitimate edge cases, including super-fans and sleep-music sessions, exhibit activity patterns that closely mimic those of coordinated fraud. We present SAGE, a novel counterfactual-aware negative harvesting approach that combines SimHash-based stratified sampling with a modular gating ensemble for confident negative identification from unlabeled data. Our ensemble architecture employs pluggable statistical gates (currently instantiated with Mahalanobis distance and k-NN density) with configurable voting thresholds enabling adaptive precision-recall trade-offs. This addresses the representation bias problem in Positive-Unlabeled learning by ensuring comprehensive coverage of rare behavioral cohorts through floor-constrained sampling. Evaluation demonstrates strong precision and recall on held-out data. The approach generalizes across fraud detection domains, achieving strong performance on both customer-level and artist-level fraud without modification to the core methodology.
\end{abstract}



\maketitle

\section{Introduction}

The music streaming industry has experienced unprecedented growth, with global recorded music revenue reaching \$29.6 billion in 2024, where streaming services account for 69\% (\$20.4 billion) of this total \cite{ifpi2025}. In the U.S., streaming represents 84\% of the market, generating over \$14 billion in revenue with nearly 97 million paid subscriptions \cite{riaa2023}. However, this financial success has attracted sophisticated fraud schemes, with investigations showing that 1-3\% of streams on major services may be fraudulent \cite{cnm2023}.

In 2024, the first-ever criminal prosecution for music streaming fraud charged a North Carolina musician with running a 7-year scheme using AI-generated songs and bot networks that generated over \$1.2 million in annual royalties \cite{doj2024}. These incidents underscore how streaming manipulation combines technological sophistication with scale to exploit royalty payment systems.

The industry has responded with increased collaboration and technological deployment. In late 2023, a coalition including Spotify, Amazon Music, and Universal Music Group formed the ``Music Fights Fraud'' task force \cite{musicafrica2024}. Music streaming services are deploying advanced AI-based detection systems at scale, with solutions processing over 2 trillion streams in 2023 \cite{beatdapp2024}, reflecting growing recognition that robust fraud detection is essential for protecting artist revenues and maintaining ecosystem integrity.

\subsection{Related Work}

\textbf{PU Learning.} Positive-Unlabeled learning \cite{elkan2008learning,liu2002partially,bekker2020learning} addresses scenarios where only positive examples are labeled. Recent advances include nnPU \cite{kiryo2017positive} and its variants \cite{chen2020positive}, which use risk estimators to handle label noise during training. However, these methods require iterating over the entire unlabeled population, which becomes computationally prohibitive at global streaming service scale.

\textbf{LSH and Sampling.} SimHash \cite{charikar2002similarity} preserves similarity through random projections, widely used for near-duplicate detection \cite{manku2007detecting} and nearest neighbor search \cite{indyk1998approximate}. We adapt SimHash for behavioral stratification to ensure rare cohorts are represented, a novel application to address representation bias in fraud detection. Our floor-constrained sampling ensures minimum representation per behavioral stratum, preventing false positives on edge cases.

\textbf{Ensemble Methods and Statistical Gates.} Ensemble approaches combine multiple models or filters to improve robustness \cite{dietterich2000ensemble}. Traditional ensembles aggregate predictions from multiple classifiers (bagging, boosting, stacking), while our gating ensemble operates at the data curation stage, filtering samples for training set construction. Mahalanobis distance \cite{mahalanobis1936generalized} with Ledoit-Wolf shrinkage \cite{ledoit2004well} provides robust multivariate outlier detection, while k-NN-based methods \cite{ramaswamy2000efficient,breunig2000lof} capture local density. Our contribution is a modular gating ensemble architecture where pluggable statistical gates vote on sample confidence for PU learning, with configurable voting thresholds enabling adaptive precision-recall trade-offs. Unlike prediction ensembles, our gates ensure samples pass both global statistical and local density checks before inclusion in training data.

\textbf{Fraud Detection.} Streaming fraud detection remains largely unexplored. Esmaeilzadeh et al. \cite{esmaeilzadeh2022abuse} use heuristic labeling for video streaming, while Sejr et al. \cite{app11052270} apply outlier detection to music data. Related work includes bot detection \cite{muralidharAAAI23} and financial fraud \cite{ngai2011application,pozzolo2015credit}. However, none address the combination of counterfactual-aware negative harvesting, global traffic scale, extreme class imbalance, and absence of systematic negative labels. These are the core challenges SAGE is designed to solve.

\section{Data and Features}

Our fraud detection system operates on customer-level behavioral data aggregated from streaming interactions over defined observation windows, formulated as binary classification distinguishing Fraud (manipulated activity) from Non-Fraud (legitimate engagement). The feature engineering process evolved through multiple iterations, guided by domain expertise and data-driven insights, converging on a representation balancing predictive power with real-time deployment constraints.

The feature space captures three critical dimensions of streaming behavior. First, we compute temporal consistency patterns through variance metrics across hourly, daily, and weekly time grains, where low variance typically indicates bot-like uniform activity while high variance suggests organic human usage. Second, we extract behavioral diversity signals including entropy and standard deviation over categorical attributes such as device types and content selection sources, where high entropy reflects varied human-like interactions and low entropy may indicate scripted behavior. Third, we compute short-term behavioral trends over a trailing observation window that prove particularly effective at distinguishing emerging fraud patterns from legitimate edge cases like super-fans or ambient listeners. This design prioritizes interpretability and computational efficiency, avoiding complex sequence models in favor of explainable statistical features that align with risk investigator intuition.

Our labeling strategy evolved significantly throughout the project life cycle. Initially, we bootstrapped training using heuristic labels derived from domain expertise and risk specialist knowledge due to the scarcity of high-quality annotated data. These heuristics partitioned the customer base into categorized and uncategorized segments, revealing extreme class imbalance with fraud cases representing approximately 1\% of the labeled population, characteristic of fraud detection problems. As our pipeline matured and we developed custom investigation tools, we transitioned to human-labeled ground truth collected through systematic manual review. This progression from heuristic to human-annotated labels enabled increasingly sophisticated modeling approaches and represents a critical evolution in our detection capabilities. The final feature set, refined through model importance scores, comprises multiple dimensions optimized for production-scale fraud detection.

\section{Approach}

\subsection{Evolution of Prior Work}
Our research progressed through multiple modeling paradigms, each addressing limitations of its predecessor. We began with unsupervised anomaly detection using Isolation Forest and Variational Autoencoders \cite{kingma2022}, which surfaced potential fraud through anomaly scores but lacked the precision required for production deployment. We then explored semi-supervised approaches including random undersampling of the majority class and student-teacher self-learning frameworks \cite{amini2024,yarowsky-1995-unsupervised}, which initially improved precision but suffered from error amplification in iterative self-labeling, leading to precision drops, along with label noise and incomplete population coverage. Cluster-based methods using K-Means for intelligent undersampling \cite{yen2009cluster} showed further gains by preserving population structure. 

However, a critical blind spot persisted across all approaches: underrepresentation of low-tenure customers and certain device cohorts in training data created systematic bias, where models flagged these segments as fraud simply due to insufficient exposure to their normal behavioral patterns. This counterfactual problem (the absence of legitimate edge cases that resemble fraud) motivated SAGE.

\subsection{Proposed Approach}

The core challenge in streaming fraud detection lies not in identifying fraud, but in comprehensively modeling legitimate behavior at scale. Traditional fraud-first approaches struggle with this, leading to elevated false positives. We adopt a "understand the haystack before looking for needles" philosophy, addressing the counterfactual problem through SAGE: a modular gating ensemble architecture combining locality-sensitive hashing for behavioral stratification with pluggable statistical gates for confident negative harvesting.

\subsubsection{SimHash-Based Stratified Sampling:} We employ SimHash \cite{charikar2002similarity}, a locality-sensitive hashing technique, to map high-dimensional customer behavior into binary signatures where behaviorally similar customers receive similar hash codes. This enables efficient stratification of millions of daily customers into thousands of behavioral buckets, ensuring proportional representation across the entire behavioral spectrum. Unlike random sampling, which underrepresents rare cohorts, our approach applies floor constraints (minimum samples per bucket) to guarantee coverage of edge cases such as super-fans and functional music listeners. SimHash-based stratification operates in O(n) time, preserving population structure while solving the representation bias problem that plagued earlier iterations. We tuned hyperparameters for this component (bucket count and floor threshold) independently to optimize behavioral coverage rather than through end-to-end grid search.

\subsubsection{Gating Ensemble for Confidence Filtering:} To harvest high-confidence negatives from the unlabeled population, we introduce a modular ensemble of statistical gates where each gate independently assesses sample confidence. The ensemble employs a configurable voting mechanism: samples are accepted based on how many gates they pass, allowing adaptive precision-recall trade-offs. For high-precision scenarios, we require unanimous voting (all gates must pass). For higher recall, we can relax to majority voting or k-out-of-n thresholds. This flexibility enables us to tune the operating point based on deployment constraints without retraining.

This ensemble architecture is extensible; additional gates such as Isolation Forest, Local Outlier Factor, or domain-specific heuristics can be plugged in without modifying the core framework. Our current instantiation employs two complementary gates fitted on the labeled fraud distribution. The first gate uses Mahalanobis distance \cite{mahalanobis1936generalized} with Ledoit-Wolf covariance shrinkage \cite{ledoit2004well} to measure global statistical distance from the fraud distribution's center, accounting for feature correlations and variance. Samples exceeding a calibrated distance threshold (far from fraud in global feature space) pass this gate. The second gate employs k-NN density estimation \cite{breunig2000lof} to capture local structure. A sample might be globally distant from fraud but locally surrounded by fraud instances, or vice versa. 

\begin{figure}[ht]
    \centering
    \includegraphics[width=0.3\textwidth]{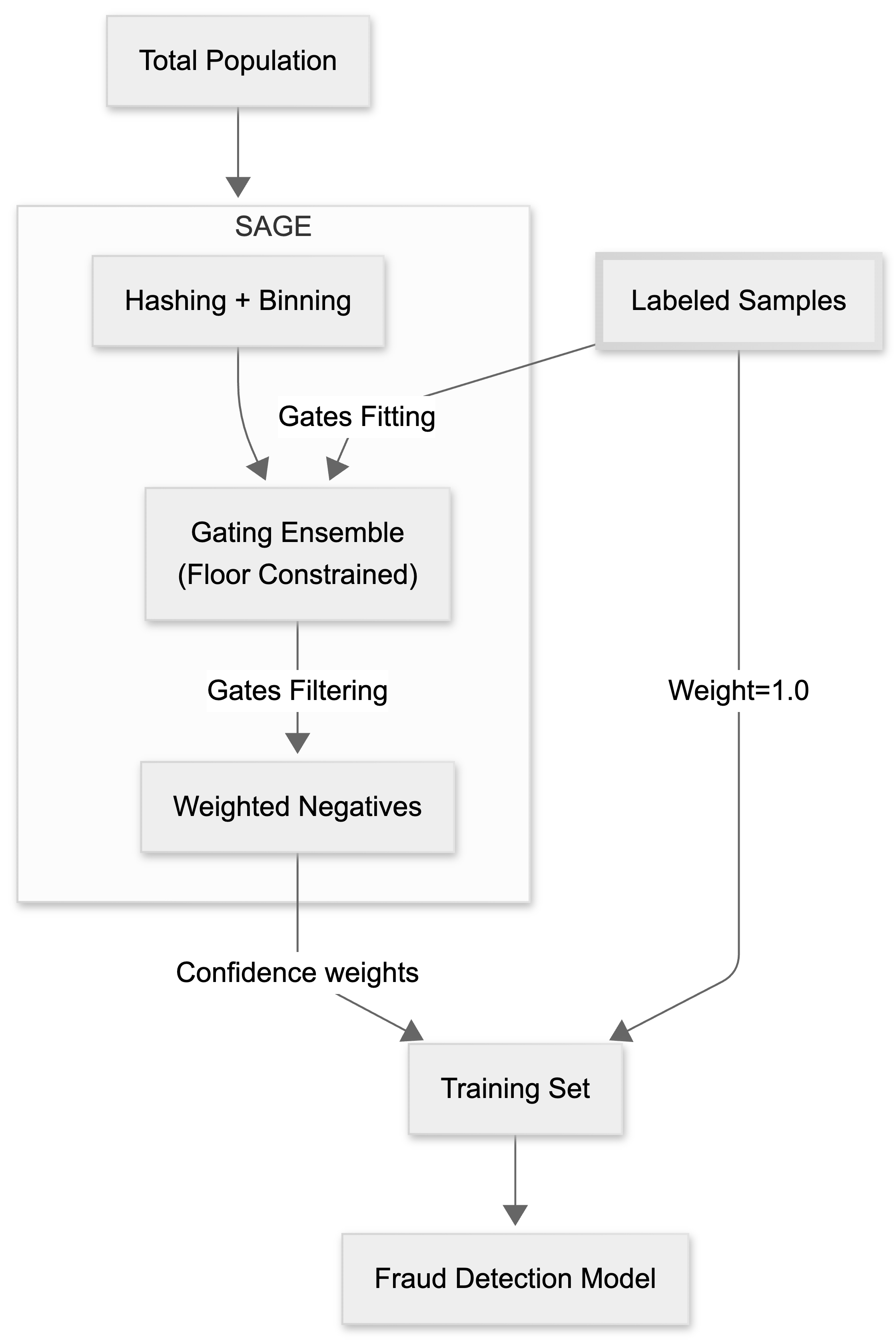}
    \caption{SAGE - SimHash stratification with floor-constrained gating ensemble to harvest confident negative samples with weights}
    \label{fig:simhash-approach}
\end{figure}

For our current model, we use unanimous voting (both gates must pass) to maximize precision in confident negative harvesting, effectively filtering contamination through complementary global and local perspectives. The dual-gate design addresses a statistical limitation: global distance alone would accept boundary samples near the fraud distribution's edge, while local density alone would reject legitimate outliers surrounded by sparse fraud instances. The combination ensures samples are both globally distant and locally dissimilar to fraud. We assign harvested negatives confidence-based weights derived from their gate scores, with samples passing gates by larger margins receiving higher weights in training. 

Threshold tuning follows a systematic procedure: we sweep candidate thresholds for each gate independently on held-out validation data, measuring contamination rates (samples incorrectly harvested as negatives). Contamination is assessed using distance-based metrics and simple regression models. Thresholds are selected to minimize contamination while maximizing negative sample yield. The dual-gate combination with unanimous voting is then validated on a separate test set to confirm low contamination is maintained.

\begin{figure}[ht]
    \centering
    \includegraphics[width=0.35\textwidth]{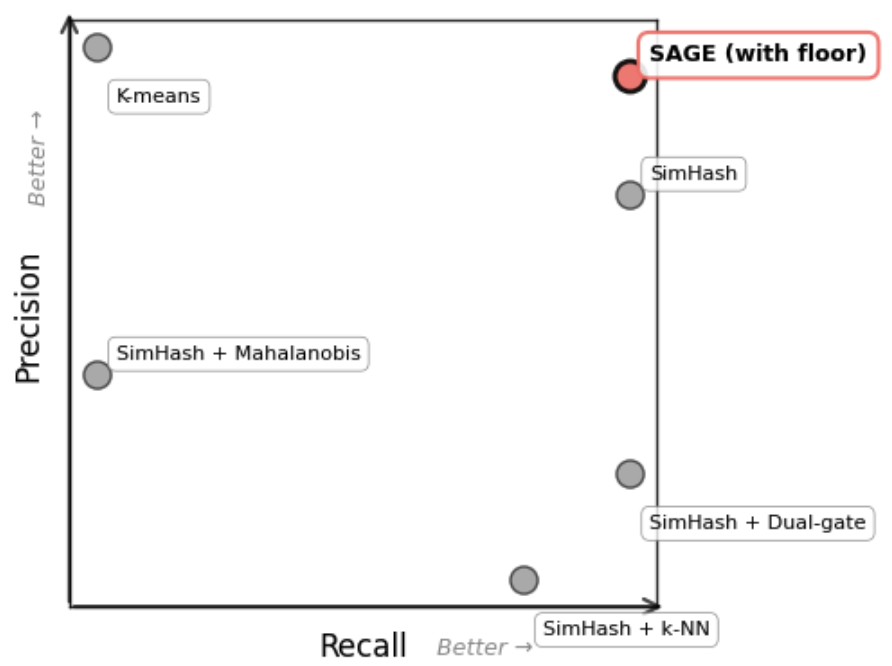}
    \caption{Precision-recall chart for ablation study comparing SimHash stratification and gating combinations}
    \label{fig:pr-ablation}
\end{figure}

To validate the effectiveness of our gating ensemble design, we conducted ablation experiments comparing different combinations of SimHash stratification and statistical gates. Figure~\ref{fig:pr-ablation} shows relative precision-recall comparison demonstrating that the combination of SimHash with dual gates (Mahalanobis + k-NN) with bin floors achieves superior performance compared to individual components or alternative configurations, confirming the complementary nature of our ensemble architecture. Figure~\ref{fig:simhash-approach} illustrates the complete pipeline from SimHash stratification through the gating ensemble to final training set construction.

\subsubsection{Multi-Class Formulation and Training:} While the core problem is binary classification (Fraud vs. Non-Fraud), human investigations revealed cases where investigators were uncertain about the fraud label. To maintain high precision and avoid forcing uncertain cases into binary labels, we introduce a third class called Suspicious and formulate the problem as multi-class classification. The foundation dataset comprises human-labeled fraud and suspicious cases from risk operations investigations, along with a small set of manually verified non-fraud samples. However, this labeled non-fraud set is insufficient to represent the full spectrum of legitimate behavior. SAGE's SimHash-gating ensemble approach addresses this by harvesting confident negatives from the unlabeled population, expanding the non-fraud set by orders of magnitude while maintaining comprehensive behavioral coverage. We train a LightGBM \cite{ke2017lightgbm} classifier on this hybrid dataset, incorporating behavioral features (temporal variance, entropy, standard deviation), ambient listening patterns, and customer attributes. The primary focus remains on the Fraud class for production decisions, while the Suspicious class provides a buffer for uncertain cases requiring manual review. This counterfactual-aware training ensures the model has seen diverse legitimate behaviors before encountering similar patterns in production, reducing false positives on edge cases.

The model operates with a two-tier decision framework. High-confidence fraud predictions trigger immediate automated annotation, while lower-confidence scores route to manual review queues. This hybrid approach balances automation with human oversight \cite{Settles2010}, achieving high precision with operationally useful recall for production deployment at scale.

\section{Results}

We evaluated our approach against multiple baseline methods, progressing from unsupervised anomaly detection through semi-supervised and self-learning approaches. Table~\ref{tab:relative_change_metrics} summarizes performance improvements relative to an Isolation Forest baseline. While prior methods achieved substantial gains, all shared a critical limitation: systematic under-representation of low-tenure customers and certain device cohorts in training data, leading to elevated false positive rates on these edge case segments.

Note that nnPU \cite{kiryo2017positive} is not included in baseline comparisons. We trained nnPU on a random undersample of unlabeled data, but this is not comparable to other approaches in this paper as it did not leverage the full unlabeled population. Training nnPU on the full unlabeled population is computationally infeasible (millions of daily customers), which is where scale becomes a critical differentiator for SAGE.

\begin{table}[h]
\centering
\caption{Relative Performance (percentage points) over baseline}
\label{tab:relative_change_metrics}
\small
\begin{tabular}{lccc}
\toprule
\textbf{Method} & \textbf{$\Delta$ Precision (pp)} & \textbf{$\Delta$ Recall (pp)} & \textbf{$\Delta$ F1 (pp)} \\
\midrule
Isolation Forest & baseline & baseline & baseline \\
Var. Auto Enc. & +55.8 & +10.1 & +19.2 \\
Random undersampling & +73.8 & \textbf{+82.2} & +78.6 \\
Student-Teacher & \textbf{+82.0} & +22.3 & +36.3 \\
Clustering (K-Means) & +80.5 & +75.0 & +78.0 \\
\textbf{SAGE (proposed)} & \textbf{+81.9} & \textbf{+87.2} & \textbf{+85.2} \\
\bottomrule
\end{tabular}
\end{table}

SAGE addresses the counterfactual problem directly and achieves the strongest overall performance, with balanced precision and recall improvements of +81.9pp and +87.2pp respectively. Evaluation on held-out data demonstrates performance across diverse customer segments, including previously problematic edge cases.

\section{Conclusion}

We present SAGE, a scalable approach for confident negative harvesting in Positive-Unlabeled learning that addresses the counterfactual problem, the absence of representative negative examples resembling positive cases. The core challenge in streaming fraud detection lies not in identifying fraud, but in comprehensively modeling legitimate behavior at scale. SAGE adopts a "understand the haystack before looking for needles" philosophy, combining SimHash-based stratified sampling with a modular gating ensemble featuring configurable voting thresholds. This enables adaptive precision-recall trade-offs without retraining. Our current model achieves +81.9pp precision and +87.2pp recall improvements over baseline, demonstrating that comprehensive legitimate behavior modeling outperforms fraud-first approaches.

The modular architecture generalizes across fraud detection domains and applies to any classification problem with scarce positive labels, absent negative labels and edge cases resembling the target class in domains like financial fraud, bot detection, spam filtering, and cybersecurity. SAGE's paradigm shift from detecting anomalies to understanding the full behavioral spectrum offers a principled path for robust classification at scale in challenging real-world scenarios.

\section{Discussion}

\textbf{Stratified Sampling Trade-offs:} Biasing learning toward long-tail and edge cases could in principle sacrifice mainstream performance. However, we observe improvements across the board through two mechanisms. First, edge-case sampling reduces label noise in the negative class by filtering contamination in behavioral overlap regions where fraud and legitimate activity are difficult to distinguish. Second, floor constraints correct under-representation without over-representation, mainstream behaviors still dominate by volume, but edge cases are no longer systematically absent. This ensures the model learns both common and rare patterns.

\textbf{Limitations:} Concept and temporal drift remain ongoing challenges inherent to the fraud detection domain, requiring regular model updates. SAGE is not immune to these shifts despite reducing false positives on edge cases. The approach requires sufficient labeled fraud samples to fit the gates, and threshold tuning demands careful validation to balance contamination and yield.

\textbf{Future Directions:} Promising extensions include incorporating temporal graph structures to capture evolving fraud networks, applying contrastive learning to enhance edge case representation, and exploring additional gates (Isolation Forest, domain-specific heuristics) within the modular ensemble.

\begin{acks}
We thank the Amazon Music Product, Engineering, and Science teams for their support and collaboration on this work. We are especially grateful to the Operations team for their domain expertise, timely feedback, and detailed manual investigations, which were instrumental in shaping the labeling strategy and validating the model outputs. This project was made possible by the collective cross‑functional effort and shared commitment to safeguarding the integrity of the Amazon Music streaming ecosystem.
\end{acks}

\bibliographystyle{ACM-Reference-Format}
\bibliography{references}

\appendix

\end{document}